# Improving Colorectal Cancer Screening and Risk Assessment through Predictive Modeling on Medical Images and Records

Shuai Jiang, PhD[1], Christina Robinson, MS[2,3], Joseph Anderson, MD[4,5], William Hisey, MSc [2,3], Lynn Butterly, MD[2,3,4], Arief Suriawinata, MD[6], Saeed Hassanpour, PhD[1,7,8,*]

[1]Department of Biomedical Data Science, Geisel School of Medicine at Dartmouth, Hanover, New Hampshire
[2]Department of Gastroenterology and Hepatology, Dartmouth-Hitchcock Medical Center, Lebanon, New Hampshire
[3]New Hampshire Colonoscopy Registry, Lebanon, New Hampshire
[4]Department of Medicine, Dartmouth-Hitchcock Medical Center, Lebanon, New Hampshire
[5]White River Junction VA Medical Center, Hartford, Vermont
[6]Department of Pathology and Laboratory Medicine, Dartmouth-Hitchcock Medical Center, Lebanon, New Hampshire
[7]Department of Epidemiology of Geisel School of Medicine at Dartmouth, Hanover, New Hampshire
[8]Department of Computer Science, Dartmouth College, Hanover, New Hampshire

* Correspondence: Saeed Hassanpour, PhD
Postal address: 1 Medical Center Drive, HB 7261, Lebanon, NH 03756
Phone: (603) 646-5715
Email: Saeed.Hassanpour@dartmouth.edu

**Conflict of interest statement:** Authors declare no conflict of interests.

**Funding Sources:** This research was supported in part by grants from the US National Library of Medicine (R01LM012837 and R01LM013833) and the US National Cancer Institute (R01CA249758).

# Abstract

Colonoscopy screening effectively identifies and removes polyps before they progress to colorectal cancer (CRC), but current follow-up guidelines rely primarily on histopathological features, overlooking other important CRC risk factors. Variability in polyp characterization among pathologists also hinders consistent surveillance decisions. Advances in digital pathology and deep learning enable the integration of pathology slides and medical records for more accurate CRC risk prediction. Using data from the New Hampshire Colonoscopy Registry, including longitudinal follow-up, we adapted a transformer-based model for histopathology image analysis to predict 5-year CRC risk. We further explored multi-modal fusion strategies to combine clinical records with deep learning-derived image features. Training the model to predict intermediate clinical variables improved 5-year CRC risk prediction (AUC = 0.630) compared to direct prediction (AUC = 0.615, p = 0.013). Incorporating both imaging and non-imaging data, without requiring manual slide review, further improved performance (AUC = 0.674) compared to traditional features from colonoscopy and microscopy reports (AUC = 0.655, p = 0.001). These results highlight the value of integrating diverse data modalities with computational methods to enhance CRC risk stratification.

# 1 Introduction

It is estimated that 52,550 lives were lost in 2023 due to colorectal cancer (CRC) in the U.S., giving CRC the highest cancer mortality rate after lung cancer. Overall, 4.3% (1 in 23) of men and 3.9% (1 in 26) of women will be diagnosed with CRC in their lifetime[1]. However, colorectal cancer can be prevented through regular screening procedures[2,3], since almost all CRC arises from colorectal polyps [thereafter, "polyp(s)"][4]. Meanwhile, nearly half of Western adults will have a polyp in their lifetime, and one-tenth of these cases will progress to cancer[5]. Fortunately, it usually takes several years for these polyps to progress, leaving a window for removal and CRC prevention.

Currently in the U.S. it is recommended that all average-risk adults undergo CRC screening by the age of 45, while patients at increased risk, such as those with a family history of CRC in a close relative, need to start screening earlier. Colonoscopy screening is associated with a 40-67% reduction in the risk of death from colorectal cancer[6,7].

There are multiple "direct- visualization" screening methods to detect polyps, such as colonoscopy, sigmoidoscopy, and virtual colonoscopy. Among those, colonoscopy has become the most common screening test in the U.S.[8]. More than 15 million colonoscopies are performed in the U.S. each year [9,10]. In 2021, it was estimated that 63.1% of US adults were up to date with colonoscopy screening, and it is expected the CRC screening will increase to cover 74.4% of Americans aged 50 to 75 years by 2030[8].

In 2020, the U.S. Multi-Society Task Force on CRC issued updated guidelines on CRC surveillance after colonoscopy screening and polypectomy[11–13]. The Task Force recognizes a range of CRC risk classifications for patients undergoing colonoscopies. The timing of the recommended follow-up colonoscopy depends on these categories: <5 years for high-risk, and 5-10 years for low-risk patients. Of note, according to the Task Force guidelines, risk and follow-up recommendations for patients undergoing colonoscopy depend only upon histopathological characterization (i.e., type), number, size, and location of polyps detected in previous colonoscopies. Therefore, the current screening and surveillance guidelines are limited and do not take into account many other CRC risk factors, such as age, race, body mass index, smoking/alcohol use, activity level, diet, and family history[14–23]. Incorporating these relevant factors in colorectal cancer risk assessment aims to capture a comprehensive context to provide patients with a more accurate and efficient surveillance plan.

Another gap in CRC risk stratification arises from the microscopic examination of polyps. Although the recommended intervals between surveillance procedures depend on histopathological analysis[12], accurate reading can be a challenging task[4] – currently, there is a significant amount of variability among pathologists in how they characterize and diagnose colorectal polyps[24–35]. For instance, sessile serrated polyps, which can develop dysplasia and progress to CRC are difficult to differentiate histologically from hyperplastic polyps[36–38]. The accurate characterization and differentiation of polyps can help patients to receive appropriate follow-up surveillance and to reduce unnecessary additional screenings, healthcare costs, and stress.

With the recent expansion of whole-slide digital scanning, high-throughput tissue banks, and archiving of digitized histological studies, the field of digital pathology is primed to benefit significantly from deep-learning models. A major advantage of deep learning for histopathological image analysis is eliminating the need to design application-specific, handcrafted features for training the model, thus can be applied to various tasks, including nucleus detection[39–41], tumor classification[42–49], and patient outcome prediction[50–53]. As transformer models reshape the landscape of medical image analysis, they have been applied to CRC-related tasks, such as CRC segmentation[54] and colorectal polyp classification[55]. In a substantial multi-center study, transformer models were observed to outperform CNN-based methods in predicting microsatellite instability[56].

Histopathological information provides crucial insights into tumor characteristics; however, for a comprehensive analysis, the inclusion of non-image features is imperative. Multi-modal fusion is a key strategy for combining information from diverse modalities to augment prediction accuracy. These fusion techniques are broadly categorized into early fusion and late fusion. Early fusion involves concatenating features from each modality and training a unified model, while late fusion entails training individual models and combining their decisions[57]. Numerous studies adopt the approach of concatenating features from different modalities to construct a unified patient representation[58–62]. Although feature-level fusion has the potential to capture intricate correlations among features, its flexibility may elevate the risk of overfitting. Conversely, some studies have demonstrated that simpler decision-level fusion techniques outperform feature-level fusion in certain scenarios[63,64].

A key requirement for developing accurate deep-learning methods is the access to large datasets for model training. Despite the rapid advancements in deep-learning technology and computational capabilities, there have not been major efforts to curate large, open access, and high-quality, annotated images for polyp analysis. The lack of "big data" in this domain makes it virtually impossible for existing or new deep-learning architectures to be developed for polyp histopathological characterization. In this study, we will utilize the data collected from New Hampshire (NH) Colonoscopy Registry (NHCR) for model development and evaluation.

The overall objective of this study is to develop and evaluate a novel, accurate, and automatic deep-learning method to analyze clinicopathological findings associated with colorectal polyps for colorectal cancer prevention by leveraging both imaging and medical record information. In a previous study[67], we developed a transformer-based pre-training and fine-tuning pipeline, MaskHIT, for medical image analysis. In this study, we adapt the MaskHIT model for automatic, accurate, and interpretable polyp characterization on histology images. In addition, we explored different fine-tuning strategies (direct vs. guided) and various multi-modal fusion techniques to improve CRC risk assessment based on both imaging and non-imaging data. We anticipate the improved accuracy in future CRC risk prediction would benefit colonoscopy participants in making more informed follow-up decisions.

**Figure 1.** Technical Overview. **a)** Illustration of the MaskHIT pipeline. b) Data sources of medical records. c) Direct prediction of 5-year CRC risk using WSI. d) Prediction of 5-year CRC risk using medical records. e) Guided attention approach that fine-tunes MaskHIT first for intermediate variables, then for the 5-year CRC risk. f) Different fusion methods.

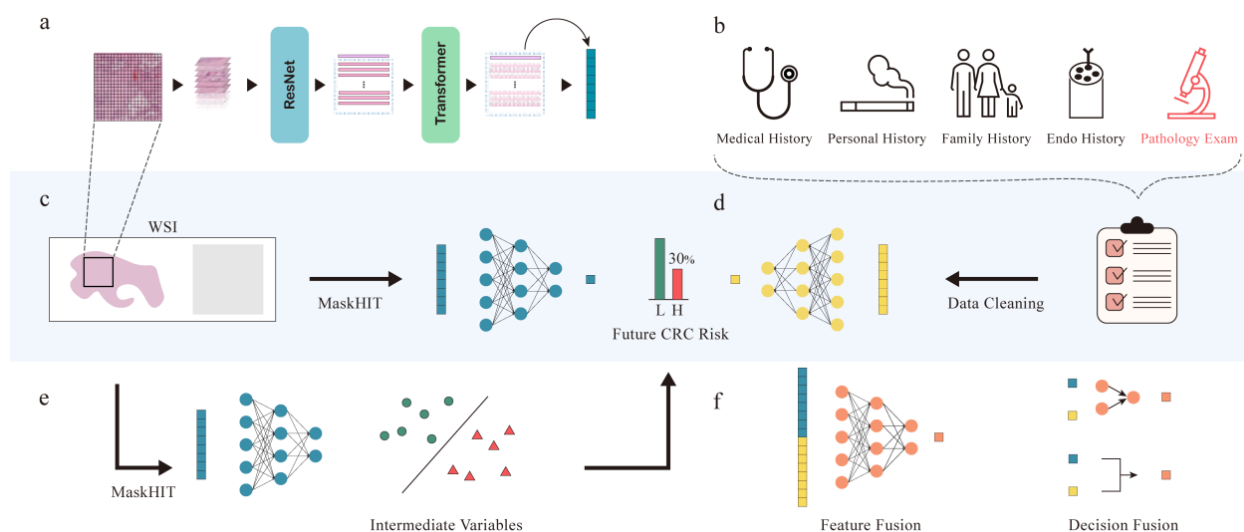

## 2 Materials and Methods

### 2.1 Study population

The NHCR is an NCI-funded, statewide registry that contains comprehensive longitudinal colonoscopy information from nearly all endoscopy sites in NH since 2004. It includes patient risk factors, such as age; sex; personal and family history of polyps or CRC; weight; height; smoking status; alcohol consumption; endoscopy history; polyp sizes, locations, numbers, and treatment; pathology reports; follow-up recommendations; and follow-up outcomes[65,66]. These data are extracted through a rigorous data collection effort from 31 participating practices, in addition to questionnaire responses from patients. The NHCR's collected data are unique in the U.S., in terms of its detailed, population-based, longitudinal and complete data. Dartmouth Hitchcock Medical Center (DHMC), a tertiary academic medical center in Lebanon, NH, has been participating in NHCR from its start in 2004 and, among patients who have their information recorded in the NHCR, over 30,000 of them are DHMC patients. The histology slides of these DHMC patients are stored at the Department of Pathology and Laboratory Medicine (DPLM) at DHMC and were available for this study.

The sources of data in this project include NHCR[68,69] and pathology slides from DHMC. A total of 2,598 patients who underwent colonoscopy at DHMC from 2004 to 2018 without a CRC diagnosis at the index visit, and had digitized polyp slides from their baseline colonoscopies with CRC status reassessed after 5 years, were included in this study. After excluding 205 patients with missing clinical data, 2,393 patients were included in the training and evaluation of our proposed models. This dataset encompasses hematoxylin and eosin (H&E) stained WSIs with various types of polyps and normal cases, along with patients' clinical records from NHCR.

## 2.2 Outcome

The NHCR and DHMC follow-up medical records are used to identify high-risk of CRC and to build the CRC risk reference standard labels for patients. Based on polyp recurrence rate, CRC progression time, and the recommended frequency for follow-up colonoscopies[11,12], in this study, we consider patients who developed CRC, advanced adenomatous polyps, or serrated polyps with dysplasia in the 5 years after screening as high-risk, while patients without those developments in 5 years following their baseline colonoscopies as low-risk. Advanced adenomatous polyps include polyps ≥1 cm, with villous components (TVA/V), or with high-grade dysplasia[70]. Advanced adenomas and serrated polyps ≥1 cm or with dysplasia are known as surrogates for CRC and are widely used as high CRC risk indicators[71–75]. The 5-year risk window is chosen to maximize the clinical utility, based on our use case in this project and the current guidelines for follow-up colonoscopy rate[11,12].

## 2.3 Clinical features and image data

Under the review and approval of the Committee for the Protection of Human Subjects, we extracted the following information from the NHCR database: 1) Identifiers of patients with tissue removed during colonoscopy, including the pathology Case-ID, used to locate tissue slides and access WSIs; 2) types, numbers, and sizes of polyps identified in the baseline colonoscopy exam; 3) microscopic determination of the tissue type of the polyps; and 4) relevant medical information for patients from the NHCR relational database.

The medical information was collected from NHCR Procedure Forms, completed by endoscopists or endoscopy nurses at participating sites and patient questionnaire responses. The NHCR database covers a comprehensive list of CRC risk factors based on peer-reviewed publications[65,66]. The variables we extracted from the NHCR database are summarized in four categories, as shown in Supplementary Table 1.

H&E-stained WSIs, scanned at DHMC (Aperio AT2, Leica Biosystems), were processed using MaskHIT pipeline. Briefly, color thresholding technique was used to create tissue masks. Non-overlapping patches of size 224μm ×224μm (i.e., 448×448 pixels) on 20× (0.5 μm/pixel) magnification level were extracted, along with their positions on the WSI.

## 2.4 Risk prediction using WSIs

The MaskHIT architecture is employed for predicting 5-year CRC risk using WSIs. MaskHIT can effectively model the relative positional information of patches on a large region from WSI. In the pre-training phase, Masked AutoEncoder (MAE) technique was used, by first randomly masking out a portion of patches from the sampled region, then using the output from the transformer model to restore the feature representations of those masked locations. This process helps the model capture relationships between different patches and their histopathologic features and understand the context. The original MaskHIT model was pretrained using more than 10 cancer types from the Cancer Genome Atlas (TCGA) database. The MaskHIT model achieved improved performance in cancer survival prediction and cancer subtype classification tasks compared to state-of-the-art models.

The workflow of the MaskHIT model involves the extraction of square regions, comprising up to 400 patches, from WSIs. Subsequently, a ResNet model, pre-trained on ImageNet data[76], is utilized for feature extraction. The location information of the patches, along with the extracted features, is then fed into a transformer model, which comprises 8 attention heads and 12 attention layers. The output of the transformer model yields a class token, serving as a representation of the entire region. Multiple regions can be sampled from a single WSI concurrently, and the class tokens are averaged to generate a global summarization of the WSI. This global summarization is then employed for risk classification through a linear projection layer (Figure 1a, Figure 1c).

To tailor the MaskHIT model pre-trained on TCGA to the specific context of polyp analysis, we conducted an additional pre-training phase for 200 epochs, employing the same pre-training methodology as previously described[77]. During the fine-tuning stage, we randomly sampled 4 non-overlapping regions from each patient. Each region comprised up to 400 patches, each of size 448×448 pixels at a magnification level of 20×. To mitigate computational costs, 25% of patches were randomly sampled from each region. During the evaluation phase, a maximum of 64 regions were sampled from each patient across all slides that this patient has, to estimate their respective CRC risk.

Beyond the direct risk prediction using WSIs, an alternative training approach, named "guided prediction" (Figure 1e), was explored. In this procedure, a MaskHIT model was initially fine-tuned to predict intermediate variables derived from patients' medical records or microscopic examinations of polyps. Subsequently, this model was employed to predict patients' future CRC risk. Two strategies were compared: 1) freezing the weights of the transformer model and exclusively fine-tuning the last linear projection layer for outcome prediction; 2) fine-tuning both the transformer model weights and the last linear projection layer. The guided prediction approach only utilizes intermediate variables during the training procedure to assist the MaskHIT model in focusing on more relevant regions in the WSIs.

## 2.5 Combination of clinical and image histopathological information

The non-imaging variables underwent preprocessing, where continuous variables were standardized, and categorical variables were one-hot encoded, resulting in a feature vector of dimension 69. Missing values were imputed by replacing them with either the average value for continuous variables or the most common class for categorical variables.

Common approaches for risk prediction, such as penalized logistic regression, multi-layer perceptron network (MLP), and random forest (RF), were explored for modeling non-imaging variables. Each approach was evaluated through cross-validation on a small subset withheld from the training data to select the optimal architecture for modeling non-imaging variables.

We excluded microscopy exam findings from building the final fusion model given the forementioned burden and variability related to obtaining such information from colonoscopy exams. Various strategies for integrating non-imaging variables with WSI features were investigated, including feature-level aggregation and decision-level aggregation (Figure 1f). Feature-level aggregation, belonging to the early fusion technique, involved concatenating non-imaging features with those extracted from the transformer output of WSIs. Subsequently, patient outcomes were predicted using multiple layers of linear projections. In decision-level fusion, predicted risk probabilities from non-imaging and WSIs were combined either through averaging or by assigning different weights to each component.

## 2.6 Evaluation

In this study, 25% of our dataset (i.e., slides and records for 600 patients) was held out as the test set for the evaluation of the developed methods, and the remaining 75% was used as the training set. A five-fold cross-validation was conducted on the training dataset for hyperparameter tuning. We use the area under the curve (AUC) of receiver operating characteristic (ROC) to assess the model's performance. To ensure a more robust estimate of model performance, the train/test splitting process was repeated 10 times, and the average performance along with the standard deviation on the test splits was reported. Paired t-tests across repeated experiments were utilized to calculate the statistical significance ($p < 0.05$) of the compared methods.

## 2.7 Model interpretation and visualization

A significant limitation of current deep-learning methods is their black-box nature, where the focus is primarily on the efficacy of the final results, with little attention given to providing clear explanations or evidence of the factors that contribute to these outcomes. To address this issue and gain a deeper understanding of the pertinent regions on WSI influencing risk predictions, we computed the difference in attention scores between the pre-trained transformer model and the transformer model fine-tuned for outcome prediction. These attention score differences were then color-coded and overlaid onto the WSIs, enabling insights into the shift in model attention for each specific outcome prediction task.

For the multi-modal fusion model that integrates non-imaging information with WSI, interpretability was enhanced by calculating Shapley values for both non-imaging features and WSI risk predictions. These Shapley values were aggregated across repeated experiments, and the average scores were plotted for visualization, providing a transparent depiction of the contributions of each feature to the model's predictions.

# 3 Results

## 3.1 Description of study population

A description of the demographical features of the study population is shown in Supplementary Table 2. Of 2,393 patients, 1,994 (83.3%) remained in the low-risk category after 5 years, while 399 (16.7%) developed high risk findings. The patients who developed high CRC risk in 5 years were significantly older than those remained in the low CRC risk category (62.0 vs. 58.7 years old, $p < 0.001$), and were more likely to be male (60.7% vs. 51.9%, $p = 0.002$). The majority of the study population are non-Hispanic Caucasian, and the distribution of race and ethnicity does not differ by risk group. The description of other groups of variables are available in the supplementary materials.

## 3.2 Risk prediction using WSIs

In the direct prediction of 5-year CRC risk using WSIs, the MaskHIT model attained an average AUC of 0.615. We evaluated multiple intermediate variables, including size and number of adenomas, size and number of serrated lesions, and all of them combined. Furthermore, the use of polyp tissue types obtained from microscopic examinations as intermediate variables was explored, including most advanced serrated lesion, and most advanced adenoma (Table 1).

MaskHIT demonstrates robust predictive performance for various intermediate variables, with notable AUC values. The highest AUC is achieved when predicting the most advanced serrated lesion (AUC: $0.927 \pm 0.007$), followed closely by predictions for the most advanced adenoma (AUC: $0.902 \pm 0.004$). The prediction of the number of adenomas found in colonoscopy yields a slightly lower AUC at $0.800 \pm 0.007$. Overall, MaskHIT exhibits effective predictive capabilities across a range of intermediate variables.

These colonoscopy or microscopy findings can predict 5-year CRC risk with various performances (Table 1). Measurements of the size and number of adenomas were better at predicting 5-year CRC risk than measurements of serrated lesions. The best predictor among them was number of adenomas (AUC: $0.643 \pm 0.029$), while the AUCs obtained using measurements of serrated lesions were no better than random guess. Microscopy measurements, while still contributing to prediction, achieve an AUC of approximately 0.55 in forecasting 5-year CRC risk.

Upon employing the guided attention approach, where the MaskHIT model was initially fine-tuned for intermediate variables and subsequently fine-tuned for 5-year CRC risk prediction, most intermediate variables exhibited an enhancement in outcome prediction performance. The best performance was observed when utilizing the size of the largest known serrated lesion as the intermediate variable, achieving an AUC of $0.629 \pm 0.016$, although this variable itself cannot predict 5-year CRC risk better than random guess. When incorporating all colonoscopy variables as intermediate variables, the MaskHIT model achieved an average AUC of 0.622 ($\pm$ 0.015) when the transformer backend was frozen. Further fine-tuning the transformer backend for risk prediction resulted in an average AUC of 0.630 ($\pm$ 0.016), representing a statistically significant improvement compared to the direct prediction approach.

When variables from the microscopic exam of the polyps were used as the intermediate variables, "Most advanced serrated" polyp significantly improved the average AUC to 0.625 ($\pm$ 0.018) with transformer fine-tuned, which was statistically better than the direct prediction approach. While for "most advanced adenoma", the differences in average AUC were not statistically significant.

**Table 1.** Comparison of AUC (standard deviation) for direct prediction versus guided attention prediction of 5-year CRC risk using WSIs.

| Fine-Tuning Strategy | Intermediate Variable | WSI →Intermediate | Intermediate →Risk | WSI →Risk | |
|---|---|---|---|---|---|
| | | | | Freeze Transformer | Fine-Tune Transformer |
| Direct prediction | None | - | - | - | 0.615 (0.016) |
| Guided by colonoscopy exam | Largest known adenoma size | 0.889 (0.004) | 0.593 (0.026) | 0.625 (0.016) | 0.626 (0.016)* |
| | Number of adenomas | 0.800 (0.007) | 0.643 (0.029) | 0.624 (0.014) | 0.625 (0.016) |
| | Largest known serrated | 0.897 (0.004) | 0.511 (0.034) | 0.623 (0.018) | 0.629 (0.016)* |
| | Number of serrated lesions | 0.864 (0.002) | 0.482 (0.037) | 0.614 (0.028) | 0.622 (0.014) |
| | All above variables | - | 0.651 (0.027) | 0.622 (0.015) | 0.630 (0.016)* |
| Guided by microscopy exam | Most advanced serrated | 0.927 (0.007) | 0.546 (0.019) | 0.625 (0.018) | 0.625 (0.018)* |
| | Most advanced adenoma | 0.902 (0.004) | 0.550 (0.024) | 0.618 (0.018) | 0.620 (0.016) |
| * p-value < 0.05; P-values were calculated as comparing the guided prediction results to the direct prediction using paired t-test. | | | | | |

## 3.3 Risk prediction using medical records

The performance comparison of L2 penalized logistic regression, random forest, and neural network (NN) models for predicting 5-year CRC risk using non-imaging variables is presented in Table 2. Among these three prediction methods, no clear winner emerges. Variables extracted from colonoscopy exams demonstrated the best performance in predicting 5-year CRC risk (AUC: 0.653-0.658), followed by personal history-related variables (AUC: 0.589-0.594). Both medical history variables and endoscopy history variables showed the capability to predict 5-year AUC with an AUC of at least 0.54. However, family history variables did not seem to contribute significantly to CRC risk prediction. The combination of the two microscopy exam variables (most advanced serrated and most advanced adenoma) predicted 5-year CRC risk with an AUC between 0.54 and 0.55.

**Table 2.** Risk prediction performance using non-imaging variables, reported as AUC (standard deviation)

| Category | L2-Logistic | Random Forest | NN |
|---|---|---|---|
| Personal history | 0.594 (0.030) | 0.589 (0.024) | 0.594 (0.029) |
| Medical history | 0.543 (0.035) | 0.526 (0.026) | 0.541 (0.033) |
| Family history | 0.510 (0.022) | 0.519 (0.020) | 0.509 (0.018) |
| Endoscopy history | 0.536 (0.025) | 0.549 (0.032) | 0.546 (0.027) |
| Colonoscopy exam | 0.657 (0.020) | 0.653 (0.027) | 0.658 (0.023) |
| Microscopy exam | 0.550 (0.025) | 0.543 (0.025) | 0.545 (0.026) |

## 3.4 Multi-modal prediction

Table 3 compares different fusion strategies, including decision level average and weighting, and the incorporation of WSI predicted risk score and WSI extracted features with non-imaging features. The results were stratified by the strategy of fine-tuning the MaskHIT model for 5-year risk prediction. When the MaskHIT model was trained using the direct prediction approach, the best performance was achieved by using the weighted average of the independent probabilities from WSIs and the non-imaging information (AUC = 0.672±0.020). With guided prediction training, the best performance was 0.675 (±0.018) using average decision fusion. On average, the decision level fusion does not only provide improved performance, but also lower variation across the 10 repeated experiments compared to feature level fusion.

In Table 4, the 5-year CRC prediction performances resulting from diverse combinations of colonoscopy and microscopy findings, WSI risk predictions, and clinical variables are presented. In this experiment we used weighted decisions to fuse WSI-based predicted probabilities with the predicted probability from non-imaging features. Using only colonoscopy findings yielded an average AUC of 0.658 (±0.023). The inclusion of microscopy findings demonstrated a marginal adverse impact on the predictive performance for 5-year CRC risk, resulting in an AUC of 0.655 (±0.021). Incorporating WSI-predicted risk scores and additional clinical features led to noteworthy improvements, presenting average AUC values of 0.669 (±0.018) and 0.669 (±0.027), respectively. The combination of both aspects of information further increased the average AUC to 0.674 (±0.021), marking a statistically significant improvement compared to utilizing colonoscopy findings alone (p=0.004) and the combination of colonoscopy and microscopy findings (p=0.001).

**Table 3.** Comparison of fusion techniques for risk prediction, reported as AUC (standard deviation). Non-imaging features include colonoscopy findings and additional medical records. **Boldface** shows the best performance of each column.

| Fusion Method | Direct | Guided |
|---|---|---|
| Decision average | 0.670 (0.014) | **0.675 (0.018)** |
| Decision weighted | **0.672 (0.020)** | 0.674 (0.021) |
| WSI decision + Non-imaging features | 0.661 (0.031) | 0.668 (0.022) |
| WSI features + Non-imaging features | 0.664 (0.024) | 0.668 (0.024) |

**Table 4.** Comparison of input modalities for risk prediction, reported as AUC performance (standard deviation). **Boldface** indicates the best performance in each column.

| Input Modalities | AUC | Accuracy | F1 | Precision | Recall |
|---|---|---|---|---|---|
| Colonoscopy | 0.658 (0.023) | 0.646 (0.025) | 0.351 (0.031) | 0.251 (0.021) | 0.585 (0.068) |
| Colonoscopy+Microscopy | 0.655 (0.021) | 0.650 (0.021) | 0.346 (0.023) | 0.250 (0.015) | 0.566 (0.061) |
| Colonoscopy+WSI | 0.669 (0.018) | **0.655 (0.019)** | 0.351 (0.026) | 0.254 (0.017) | 0.571 (0.066) |
| Colonoscopy+Clinical | 0.669 (0.027) | 0.645 (0.026) | 0.359 (0.029) | 0.255 (0.023) | 0.604 (0.046) |
| Colonoscopy+WSI+Clinical | **0.674 (0.021)** | 0.644 (0.021) | **0.362 (0.027)** | **0.257 (0.019)** | **0.615 (0.059)** |

## 3.5 Model interpretation

### 3.5.1 Attention map visualization

The attention maps obtained from the MaskHIT model for two representative WSIs are presented in Figure 2a for a high-risk patient and Figure 2b for a low-risk patient. The visualization reveals that the MaskHIT model tends to focus more on the structures of polyps within the WSIs. Interestingly, the high attention areas appear similar whether guided fine-tuning methods were employed or not.

The intensity of attention weights from the direct prediction approach and the guided prediction approach was further examined by calculating the difference in attention weights between these two methods. The results are presented in panels 7 and 8 of Figure 2. The redder color in these panels indicates that the highlighted region received higher attention from the guided prediction model compared to the direct prediction model. This visualization demonstrates that the regions receiving higher attention from the guided prediction model generally align with the regions attended by both the direct prediction model and the guided prediction model. In essence, the guided prediction model exhibited greater confidence in assigning weights to regions that were deemed important for risk prediction.

**Figure 2.** Attention maps for two WSIs. a) High risk patient; b) Low risk patient. Sub panels: 1&2: slide/region thumbnails; 3&4: attention map generated from direct prediction model; 5&6: attention map generated from guided attention model; 7&8: attention maps generated by taking the difference between guided attention map and direct attention map; 9&10: detailed patch view.

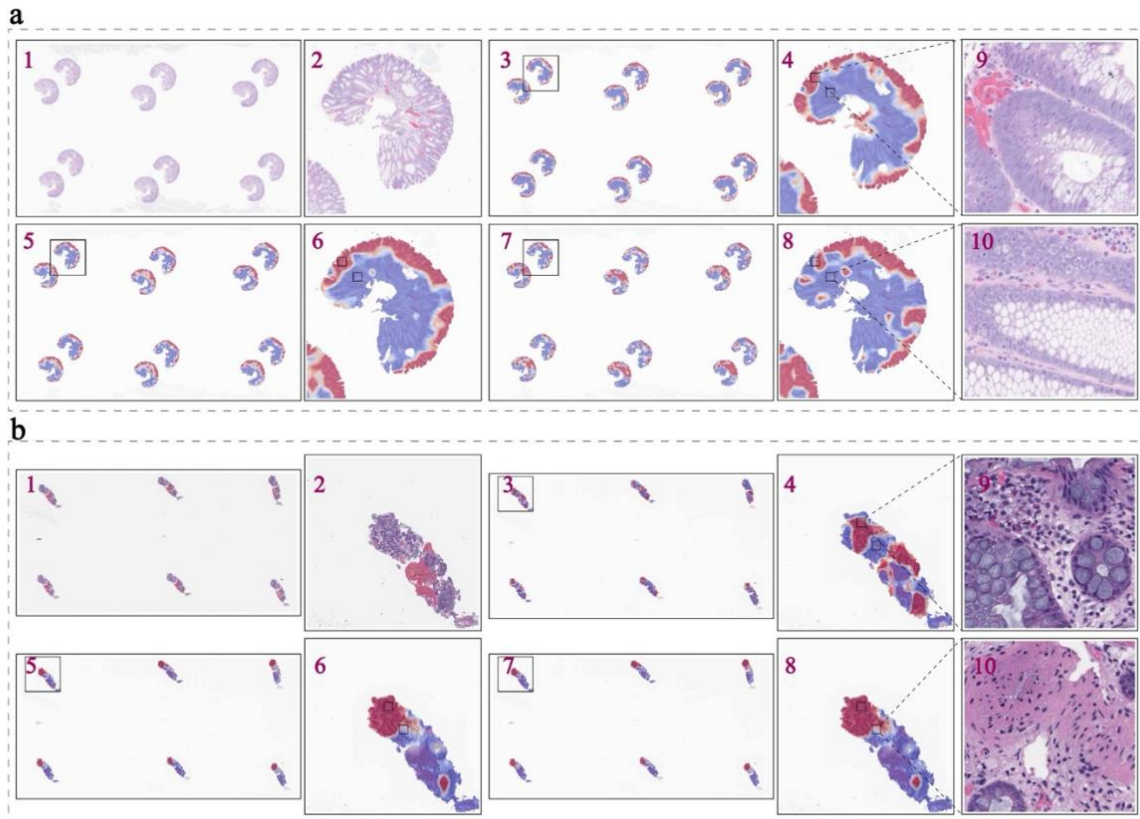

### 3.5.2 Feature importance ranking

In Figure 3, the top 10 most important features influencing the output of the final fusion model are presented. The most influential feature is the number of adenomas, demonstrating a positive association with 5-year CRC risk. Notably, the predicted risk probability from the WSI was ranked as the third most important feature in the fusion model, exceeded only by the number of adenomas and age.

**Figure 3.** Shapley values of top 10 predictors in the fusion model (CSSP: Colorectal Sessile Serrated Polyp, HP: Hyperplastic Polyp)

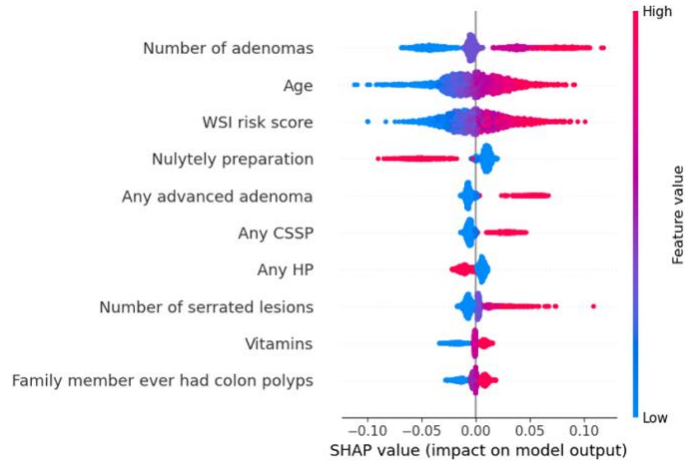

# 4 Discussion

The accurate prediction of future CRC risk is crucial for informed decisions regarding follow-up colonoscopy visits. Existing guidelines recommend leveraging polyp characteristics identified in colonoscopy exams, as well as some personal and family history risk factors, for patient risk stratification to determine the timing of subsequent colonoscopies[11]. This study sought to advance future CRC risk prediction by integrating automatic deep learning-based analysis of whole-slide images and incorporating CRC-related medical information in a predictive multi-modal pipeline.

Relying exclusively on colonoscopy and microscopy aspects resulted in an average AUC of 0.655. However, by incorporating deep learning predicted probabilities and information from clinical variables, a statistically significant improvement in the prediction AUC to 0.674 was observed. This underscores the potential of leveraging advanced computational techniques and multi-modal data fusion to enhance CRC risk assessment beyond conventional guidelines. Such an approach provides a more robust foundation for personalized and effective follow-up strategies in clinical practice.

To enhance the prediction performance utilizing WSIs, we refined our approach by adopting the recently developed model, MaskHIT. MaskHIT is a transformer-based method that leverages the location information of patches extracted from the entire slide. The unique aspect of the transformer model as a patch-level feature fusion technique lies in its capacity to incorporate spatial details, enabling our deep learning model to capture high-level structural information of the polyps. This approach stands in contrast to commonly used multiple instance learning approaches, offering a more nuanced and comprehensive representation of the intricate characteristics of colorectal polyps in our predictive model.

In addition, we conducted experiments involving a guided prediction approach to improve the transformer model for 5-year CRC risk prediction. As presented in Table 1, predicting 5-year CRC risk using WSI poses challenges due to numerous factors beyond current histopathological features from colonoscopy exams that can influence future CRC risk. Consequently, the MaskHIT model may face difficulties in confidently identifying visual features linked to CRC risk in this complex context.

To tackle this challenge, we adopted a guided prediction approach, enabling the transformer model to first predict histopathological features derived from the colonoscopy exam. Remarkably, MaskHIT demonstrated strong performance in this task, with AUCs exceeding 0.8 and considerably smaller standard deviations compared to risk prediction tasks. Subsequently, fine-tuning the MaskHIT model for risk prediction led to a statistically significant improvement compared to the direct prediction approach. Interestingly, certain variables, although ineffective at predicting future CRC risk independently, also contributed to enhancing MaskHIT's accuracy in 5-year CRC risk prediction.

Attention map visualizations supported our hypothesis, revealing that the guided prediction model assigned greater attention weights to locations relevant for risk prediction (i.e., polyps) compared to the direct prediction model. This nuanced approach demonstrates the effectiveness of leveraging the guided prediction approach to enhance the interpretability and performance of deep learning models in the context of 5-year CRC risk prediction from whole-slide images.

We conducted further exploration of various approaches for combining information from colonoscopy exams, WSIs, and clinical records. In general, decision-level fusion produced superior results compared to models combining non-image features with risk predictions from the slides. Due to the high predictive value of colonoscopy variables, the signal from WSIs predictions can be easily overwhelmed by noise in clinical features, a known issue in multi-modal fusion[78]. However, through the application of decision-level fusion techniques, this challenge can be addressed, resulting in improved outcomes compared to using either modality in isolation, consistent with findings in previous studies[63,64].

As the future steps, we plan to validate our multi-modal CRC risk approach using additional datasets. Furthermore, as our goal is to incorporate the additional information in the risk stratification for patients undergoing colonoscopy in the screening and surveillance program, we intend to quantify its health outcomes and cost impacts through follow-up clinical trials and prospective studies.

# 5 Conclusion

In this study, the integration of the transformer-predicted risk score and additional clinical information resulted in an improvement in the performance of CRC risk stratification. Notably, variables describing colonoscopy and microscopy findings of polyps were identified as contributors to enhanced performance in predicting 5-year CRC risk using deep learning models. Despite its simplicity in multi-modal fusion, decision-level fusion demonstrated superior performance improvements when combining imaging and non-imaging information. Future research is essential to refine deep learning methods for including more related clinical information and to evaluate the additional benefits of an accurate CRC risk stratification in colonoscopy screening programs.

20. Rafter, J. *et al.* Dietary synbiotics reduce cancer risk factors in polypectomized and colon cancer patients. *The American journal of clinical nutrition* **85**, 488–96 (2007).

21. Benito, E. *et al.* Nutritional factors in colorectal cancer risk: a case-control study in Majorca. *International journal of cancer* **49**, 161–7 (1991).

22. Fedirko, V. *et al.* Alcohol drinking and colorectal cancer risk: an overall and dose-response meta-analysis of published studies. *Annals of Oncology* **22**, 1958–1972 (2011).

23. Giovannucci, E. An updated review of the epidemiological evidence that cigarette smoking increases risk of colorectal cancer. *Cancer epidemiology, biomarkers & prevention : a publication of the American Association for Cancer Research, cosponsored by the American Society of Preventive Oncology* **10**, 725–31 (2001).

24. Rex, D. K., Alikhan, M., Cummings, O. & Ulbright, T. M. Accuracy of pathologic interpretation of colorectal polyps by general pathologists in community practice. *Gastrointestinal endoscopy* **50**, 468–74 (1999).

25. Terry, M. B. *et al.* Reliability in the classification of advanced colorectal adenomas. *Cancer epidemiology, biomarkers & prevention : a publication of the American Association for Cancer Research, cosponsored by the American Society of Preventive Oncology* **11**, 660–3 (2002).

26. Hetzel, J. T. *et al.* Variation in the Detection of Serrated Polyps in an Average Risk Colorectal Cancer Screening Cohort. *The American Journal of Gastroenterology* **105**, 2656–2664 (2010).

27. Jensen, P. *et al.* Observer variability in the assessment of type and dysplasia of colorectal adenomas, analyzed using kappa statistics. *Diseases of the colon and rectum* **38**, 195–8 (1995).
17

# Supplementary Materials

**Supplementary Table 1.** Extracted data from NHCR for patients undergoing colorectal polyp biopsy organized in four categories.

| Category | Extracted Data |
|---|---|
| Personal history | Age, sex, race and ethnicity, marital status |
| | Height, weight, body mass index |
| | Smoking status (duration, cigarettes per day) |
| | Alcohol consumption (drinks per week) |
| | Exercise routine, activity level |
| | Vitamin supplement use (quantity) |
| | Calcium supplement use (quantity) |
| Medical history | Familial adenomatous polyposis |
| | Hereditary non-polyposis CRC |
| | Crohn's disease or ulcerative colitis |
| | Constipation or colorectal bleeding |
| | Aspirin use (dosage) |
| Family history | Relatives diagnosed with CRC (mother/father/sister/brother/child) |
| | Age of relatives at diagnosis (<50, 50-60, >60) |
| | Relatives diagnosed with colorectal polyps, familial adenomatous polyposis, hereditary non-polyposis CRC, familial polyposis |
| Endoscopy history | Preparation type |
| | Last sigmoidoscopy and colonoscopy time |
| | Last sigmoidoscopy and colonoscopy outcome |

**Supplementary Table 1.** Demographical features of the study population

| Variable | Level | Missing | Grouped by risk | | P-Value |
|---|---|---|---|---|---|
| | | | Low Risk | High Risk | |
| n | | | 1994 | 399 | |
| Age, mean (SD) | | 0 | 58.7 (9.8) | 62.0 (8.5) | <0.001 |
| Sex, n (%) | F | 0 | 959 (48.1) | 157 (39.3) | 0.002 |
| | M | | 1035 (51.9) | 242 (60.7) | |
| Hispanic, n (%) | Hispanic | 456 | 18 (1.1) | 4 (1.2) | 0.778 |
| | Not Hispanic | | 1593 (98.9) | 322 (98.8) | |
| Race, n (%) | African American | 454 | 5 (0.3) | | 0.509 |
| | American Indian | | 3 (0.2) | | |
| | Asian | | 6 (0.4) | 2 (0.6) | |
| | Caucasian | | 1548 (95.9) | 314 (96.6) | |
| | Multiple | | 41 (2.5) | 9 (2.8) | |
| | Other | | 11 (0.7) | | |

**Supplementary Table 3.** Patient description: Personal history

| Variable | Level | Missing | Grouped by risk | | P-Value |
| --- | --- | --- | --- | --- | --- |
| | | | Low Risk | High Risk | |
| n | | | 1994 | 399 | |
| Age, mean (SD) | | 0 | 58.7 (9.8) | 62.0 (8.5) | <0.001 |
| Sex, n (%) | F | 0 | 959 (48.1) | 157 (39.3) | 0.002 |
| | M | | 1035 (51.9) | 242 (60.7) | |
| Marital Status, n (%) | Single | 494 | 113 (7.2) | 20 (6.2) | 0.451 |
| | Married | | 1196 (75.8) | 241 (75.1) | |
| | Separated | | 14 (0.9) | 4 (1.2) | |
| | Divorced | | 158 (10.0) | 31 (9.7) | |
| | Widowed | | 53 (3.4) | 18 (5.6) | |
| | Living as married | | 44 (2.8) | 7 (2.2) | |
| Hispanic, n (%) | Hispanic | 456 | 18 (1.1) | 4 (1.2) | 0.778 |
| | Not Hispanic | | 1593 (98.9) | 322 (98.8) | |
| Race, n (%) | African American | 454 | 5 (0.3) | | 0.509 |
| | American Indian | | 3 (0.2) | | |
| | Asian | | 6 (0.4) | 2 (0.6) | |
| | Caucasian | | 1548 (95.9) | 314 (96.6) | |
| | Multiple | | 41 (2.5) | 9 (2.8) | |
| | Other | | 11 (0.7) | | |
| Exercise, n (%) | No exercise | 466 | 136 (8.5) | 37 (11.4) | 0.029 |
| | Active daily life | | 549 (34.2) | 128 (39.5) | |
| | 1-5 times/week | | 760 (47.4) | 126 (38.9) | |
| | 5+ times/week | | 158 (9.9) | 33 (10.2) | |
| Smoker status, n (%) | Never smoker | 453 | 744 (46.2) | 148 (45.1) | 0.905 |
| | Former smoker | | 694 (43.1) | 141 (43.0) | |
| | Current smoker | | 167 (10.4) | 38 (11.6) | |
| | Error | | 7 (0.4) | 1 (0.3) | |
| Qty smoke, n (%) | Nonsmoker | 506 | 744 (47.5) | 148 (46.1) | 0.949 |
| | 10 or fewer cigarettes/day | | 262 (16.7) | 54 (16.8) | |
| | 11-20/day | | 345 (22.0) | 71 (22.1) | |
| | 21-30/day | | 152 (9.7) | 32 (10.0) | |
| | 31+/day | | 63 (4.0) | 16 (5.0) | |
| Years of smoking, mean (SD) | | 498 | 10.3 (13.2) | 11.7 (14.8) | 0.123 |
| Weekly alcohol intake, n (%) | 0 alcoholic drinks/week | 471 | 586 (36.7) | 116 (35.6) | 0.423 |
| | 1-4 alcoholic drinks/week | | 504 (31.6) | 101 (31.0) | |
| | 5-8 alcoholic drinks/week | | 250 (15.7) | 51 (15.6) | |
| | 9-20 alcoholic drinks/week | | 232 (14.5) | 48 (14.7) | |
| | 21+ alcoholic drinks/week | | 24 (1.5) | 10 (3.1) | |
| Calcium, n (%) | No calcium use | 1854 | 292 (61.5) | 43 (67.2) | 0.455 |
| | Calcium use | | 183 (38.5) | 21 (32.8) | |
| Vitamins, n (%) | No vitamin use | 1012 | 383 (34.2) | 76 (29.2) | 0.147 |
| | Vitamin use | | 738 (65.8) | 184 (70.8) | |
| Patient weight in inches, mean (SD) | | 462 | 183.1 (42.4) | 191.2 (43.2) | 0.002 |
| Patient height in pounds, mean (SD) | | 459 | 67.2 (4.2) | 67.8 (4.5) | 0.015 |
| BMI, median [Q1, Q3] | | 482 | 27.0 [24.0,31.0] | 28.0 [25.0,32.0] | 0.014 |

**Supplementary Table 4.** Patient description: Medical history

| Variable | Level | Missing | Grouped by risk | | P-Value |
| --- | --- | --- | --- | --- | --- |
| | | | Low risk | High risk | |
| n | | | 1994 | 399 | |
| History of IBD, n (%) | No | 152 | 1764 (94.3) | 357 (96.5) | 0.111 |
| | Yes | | 107 (5.7) | 13 (3.5) | |
| Genetic syndrome, n (%) | No | 336 | 1711 (99.9) | 344 (100.0) | 1.000 |
| | Yes | | 2 (0.1) | | |
| Diag exam-change in bowel habits, n (%) | No | 351 | 1691 (99.1) | 335 (99.7) | 0.495 |
| | Yes | | 15 (0.9) | 1 (0.3) | |
| Diag exam-Evaluate GI bleeding, n (%) | No | 351 | 1659 (97.2) | 325 (96.7) | 0.731 |
| | Yes | | 47 (2.8) | 11 (3.3) | |
| Aspirin use, n (%) | No | 527 | 896 (57.5) | 150 (48.5) | 0.004 |
| | Yes | | 661 (42.5) | 159 (51.5) | |

**Supplementary Table 5.** Patient description: Family history

| Variable | Level | Missing | Grouped by risk | | P-Value |
| --- | --- | --- | --- | --- | --- |
| | | | Low risk | High risk | |
| n | | | 1994 | 399 | |
| Screening exam for family hx of polyp(s), n (%) | No | 351 | 1584 (92.8) | 315 (93.8) | 0.635 |
| | Yes | | 122 (7.2) | 21 (6.2) | |
| Has a biological family member ever had colon polyps, n (%) | No | 447 | 366 (22.6) | 59 (18.0) | 0.129 |
| | Yes | | 522 (32.3) | 105 (32.0) | |
| | Don't know | | 730 (45.1) | 164 (50.0) | |
| Has a family history of CRC in a first-degree relative, n (%) | No | 417 | 1196 (73.5) | 253 (72.7) | 0.822 |
| | Yes | | 432 (26.5) | 95 (27.3) | |
| Has a family history of CRC in first-degree relative under 50, n (%) | No | 641 | 1369 (94.3) | 279 (93.0) | 0.470 |
| | Yes | | 83 (5.7) | 21 (7.0) | |
| Has a family history of CRC in first-degree relative under 60, n (%) | No | 660 | 1276 (89.4) | 264 (86.6) | 0.190 |
| | Yes | | 152 (10.6) | 41 (13.4) | |
| Patient family member has genetic syndrome, n (%) | No | 344 | 1684 (98.7) | 338 (98.5) | 0.795 |
| | Yes | | 22 (1.3) | 5 (1.5) | |

**Supplementary Table 6.** Patient description: Endoscopy history

| Variable | Level | Missing | Grouped by risk | | P-Value |
|---|---|---|---|---|---|
| | | | Low risk | High risk | |
| n | | | 1994 | 399 | |
| Procedure - exam preparation quality, n (%) | Excellent | 565 | 486 (31.7) | 74 (25.1) | 0.006 |
| | Good | | 906 (59.1) | 203 (68.8) | |
| | Fair | | 141 (9.2) | 18 (6.1) | |
| Type of preparation: Nulytely, n (%) | No | 104 | 1567 (81.8) | 341 (91.2) | <0.001 |
| | Yes | | 348 (18.2) | 33 (8.8) | |
| Type of preparation: Osmoprep (pills), n (%) | No | 104 | 1915 (100.0) | 374 (100.0) | 1.000 |
| Type of preparation: Half Lytely, n (%) | No | 104 | 1866 (97.4) | 368 (98.4) | 0.359 |
| | Yes | | 49 (2.6) | 6 (1.6) | |
| Type of preparation: Fleet, n (%) | No | 104 | 1909 (99.7) | 373 (99.7) | 1.000 |
| | Yes | | 6 (0.3) | 1 (0.3) | |
| Type of preparation: Other, n (%) | No | 104 | 1907 (99.6) | 372 (99.5) | 0.672 |
| | Yes | | 8 (0.4) | 2 (0.5) | |
| Last sigmoid/colonoscopy Pilot only, n (%) | Never | 1835 | 141 (28.7) | 17 (25.8) | 0.533 |
| | Within last 12 months | | 12 (2.4) | 4 (6.1) | |
| | 1-4 years ago | | 114 (23.2) | 17 (25.8) | |
| | 5-10 years ago | | 202 (41.1) | 25 (37.9) | |
| | More than 10 years ago | | 23 (4.7) | 3 (4.5) | |
| Time since last colonoscopy, n (%) | Never | 982 | 381 (33.1) | 76 (29.2) | 0.254 |
| | Within last 12 months | | 39 (3.4) | 10 (3.8) | |
| | 1-4 years ago | | 211 (18.2) | 44 (16.9) | |
| | 5-10 years ago | | 464 (40.3) | 108 (41.5) | |
| | More than 10 years ago | | 56 (4.9) | 22 (8.5) | |
| Previous colonoscopy findings: polyps, n (%) | No | 979 | 760 (65.9) | 155 (59.4) | 0.055 |
| | Yes | | 393 (34.1) | 106 (40.6) | |
| Previous colonoscopy findings: diverticulosis, n (%) | No | 979 | 1087 (94.3) | 248 (95.0) | 0.747 |
| | Yes | | 66 (5.7) | 13 (5.0) | |
| Previous colonoscopy findings: hemorrhoids, n (%) | No | 979 | 1081 (93.8) | 241 (92.3) | 0.484 |
| | Yes | | 72 (6.2) | 20 (7.7) | |
| Previous colonoscopy findings: other, n (%) | No | 979 | 1108 (96.1) | 253 (96.9) | 0.643 |
| | Yes | | 45 (3.9) | 8 (3.1) | |
| Previous colonoscopy findings: no findings, n (%) | No | 979 | 842 (73.0) | 193 (73.9) | 0.822 |
| | Yes | | 311 (27.0) | 68 (26.1) | |
| Previous colonoscopy results: do not know, n (%) | No | 1872 | 386 (96.0) | 115 (96.6) | 1.000 |
| | Yes | | 16 (4.0) | 4 (3.4) | |

**Supplementary Table 7.** Patient description: Current colonoscopy exam

| Variable | Level | Missing | Grouped by risk | | P-Value |
| --- | --- | --- | --- | --- | --- |
| | | | Low risk | High risk | |
| n | | | 1994 | 399 | |
| Number of records where an adenoma was indicated, median [Q1, Q3] | | 0 | 1.0 [0.0,1.0] | 1.0 [1.0,3.0] | <0.001 |
| Largest known adenoma size, n (%) | No adenoma | 148 | 599 (31.8) | 70 (19.2) | <0.001 |
| | <5mm | | 788 (41.9) | 142 (39.0) | |
| | 5-9mm | | 333 (17.7) | 96 (26.4) | |
| | 10-20mm | | 132 (7.0) | 50 (13.7) | |
| | >20mm | | 29 (1.5) | 6 (1.6) | |
| Any advanced adenoma at the procedure level, n (%) | No | 186 | 1657 (89.3) | 273 (77.8) | <0.001 |
| | Yes | | 199 (10.7) | 78 (22.2) | |
| Any adenoma at the procedure level, n (%) | No | 0 | 599 (30.0) | 70 (17.5) | <0.001 |
| | Yes | | 1395 (70.0) | 329 (82.5) | |
| Any HP at the procedure level, n (%) | No | 0 | 1182 (59.3) | 274 (68.7) | 0.001 |
| | Yes | | 812 (40.7) | 125 (31.3) | |
| Any SSA/P or TSA w/wo dysplasia at the procedure level, n (%) | No | 0 | 1784 (89.5) | 342 (85.7) | 0.037 |
| | Yes | | 210 (10.5) | 57 (14.3) | |
| Any CSSP at the procedure level, n (%) | No | 164 | 1569 (84.9) | 306 (80.5) | 0.043 |
| | Yes | | 280 (15.1) | 74 (19.5) | |
| Number of serrated lesions identified, median [Q1, Q3] | | 0 | 0.0 [0.0,1.0] | 0.0 [0.0,1.0] | 0.083 |
| Largest known serrated (missing path/size ignored), n (%) | No serrated polyp | 140 | 1036 (55.3) | 238 (62.5) | <0.001 |
| | <5mm | | 585 (31.2) | 75 (19.7) | |
| | 5-9mm | | 171 (9.1) | 48 (12.6) | |
| | 10-20mm | | 71 (3.8) | 17 (4.5) | |
| | >20mm | | 9 (0.5) | 3 (0.8) | |

**Supplementary Table 8.** Patient description: Pathologist assessments

| Variable | Level | Missing | Grouped by risk | | P-Value |
| --- | --- | --- | --- | --- | --- |
| | | | Low risk | High risk | |
| n | | | 1994 | 399 | |
| Most advanced adenoma at tissue level, n (%) | No adenoma | 0 | 781 (39.2) | 119 (29.8) | 0.002 |
| | Tubular adenoma | | 1159 (58.1) | 263 (65.9) | |
| | Tubulovillous adenoma | | 48 (2.4) | 14 (3.5) | |
| | Villous adenoma | | 6 (0.3) | 3 (0.8) | |
| Most advanced serrated at tissue level, n (%) | No serrated polyp | 0 | 1353 (67.9) | 299 (74.9) | 0.001 |
| | Hyperplastic polyp | | 489 (24.5) | 61 (15.3) | |
| | SSP without dysplasia | | 129 (6.5) | 35 (8.8) | |
| | SSP with dysplasia | | 15 (0.8) | 2 (0.5) | |
| | TSA | | 8 (0.4) | 2 (0.5) | |

**Supplementary Table 9.** Feature importance calculated from permutation

| Variable | Importance |
| --- | --- |
| Number of adenomas | 0.029 (0.010) |
| WSI risk score | 0.016 (0.009) |
| Age | 0.016 (0.002) |
| Nulytely preparation | 0.011 (0.004) |
| Any CSSP | 0.006 (0.002) |
| Number of serrated lesions | 0.005 (0.003) |
| Any SSA/P or TSA w/wo dysplasia | 0.004 (0.001) |
| Any advanced adenoma | 0.003 (0.002) |
| Vitamins | 0.003 (0.002) |
| Family member ever had colon polyps | 0.002 (0.001) |